\documentclass[conference]{IEEEtran}

\IEEEoverridecommandlockouts
\usepackage[shortlabels]{enumitem}
\usepackage{cite}
\usepackage{amsmath,amssymb,amsfonts}
\usepackage{algorithmic}
\usepackage{graphicx}
\usepackage{textcomp}
\usepackage{bm}
\usepackage{flushend}
\usepackage{hyperref}
\usepackage{rotating}
\usepackage{tikz}
\usepackage{booktabs,tabularx}
\usepackage[caption=false,font=normalsize,labelfont=sf,textfont=sf]{subfig}
\def\BibTeX{{\rm B\kern-.05em{\sc i\kern-.025em b}\kern-.08em
    T\kern-.1667em\lower.7ex\hbox{E}\kern-.125emX}}

\begin{document}

\title{Heartbeat Anomaly Detection\\using Adversarial Oversampling}

\author{
\IEEEauthorblockN{Jefferson L. P. Lima}
\IEEEauthorblockA{\textit{Centro de Inform\'{a}tica} \\
\textit{Universidade Federal de Pernambuco}\\
50.740-560, Recife, PE, Brazil\\
jlpl@cin.ufpe.br}\\
\and
\IEEEauthorblockN{David Mac\^edo}
\IEEEauthorblockA{\textit{Centro de Inform\'{a}tica} \\
\textit{Universidade Federal de Pernambuco}\\
50.740-560, Recife, PE, Brazil\\
dlm@cin.ufpe.br}
\and
\IEEEauthorblockN{Cleber Zanchettin}
\IEEEauthorblockA{\textit{Centro de Inform\'{a}tica} \\
\textit{Universidade Federal de Pernambuco}\\
50.740-560, Recife, PE, Brazil\\
cz@cin.ufpe.br}
}

\maketitle

\begin{abstract}

Cardiovascular diseases are one of the most common causes of death in the world. Prevention, knowledge of previous cases in the family, and early detection is the best strategy to reduce this fact. Different machine learning approaches to automatic diagnostic are being proposed to this task. As in most health problems, the imbalance between examples and classes is predominant in this problem and affects the performance of the automated solution.  In this paper, we address the classification of heartbeats images in different cardiovascular diseases. We propose a two-dimensional Convolutional Neural Network for classification after using a \textit{InfoGAN} architecture for generating synthetic images to unbalanced classes. We call this proposal Adversarial Oversampling and compare it with the classical oversampling methods as SMOTE, ADASYN, and RandomOversampling. The results show that the proposed approach improves the classifier performance for the minority classes without harming the performance in the balanced classes. 

\end{abstract}


\section{Introduction}

Cardiovascular diseases are demanding attention since they are among the leading causes of death in the world. Cardiac arrhythmia consists of disturbances that alter heart rate, constituting a severe public health problem. The electrocardiogram (ECG) is the process of recording the heart electrical activity over some time using electrodes. ECG can be handy to assist the cardiac arrhythmia diagnostic as it is a non-invasive method of detecting abnormal cadences of the heartbeats.

Cardiologists often perform human arrhythmia detection using ECG because of the high error rates of the computerized approaches. In order to perform arrhythmias detection using ECG, a machine learning algorithm must recognize the distinct arrhythmias wave types and its distinct forms. Detect arrhythmias is a difficult task due to the variability in wave morphology between patients as well as the presence of noise. The considerable intra-class variation makes the challenger even harder.

Most of the current approaches to deal with arrhythmia classification make use of feature extraction procedures. However, these approaches require a prior and specific knowledge about the characteristics that are essential to represent a specific type of heartbeat. A possible strategy is taking abnormal beats as crucial points and analyzing the recurrence of the anomalies along the segment of the ECG.
To overcome the previously mentioned limitations, considering the difficulty and the importance of taking spatial information in ECG analysis, Convolutional Neural Networks (CNN) have been used to tackle this problem in several studies \cite{1} \cite{2} \cite{3} \cite{4}.

However, one problem commonly encountered in datasets involving health is the high degree of imbalance.  Often, we come across a smaller amount of samples involving the positive class (representing the disease).  Many learning systems are not used to deal with unbalanced data. 

The use of this type of datasets to train models such as a neural network usually causes high accuracy for data belonging to the majority class, but an unacceptable accuracy for minority class data. It typically happens because the model ends up specializing only in the classes with the most significant quantity of samples as the classes with few samples are presented fewer times for the adjustments of the weights.

Many techniques try to increase the amount of minority class samples. The simplest way is to repeat samples randomly. Other techniques try to use some criterion to select the samples, such as a level of difficulty that they have according to the classification model. Some more elaborate techniques to generate extra samples trying to follow the original distribution of the data, however, much class noise is inserted in the dataset since samples may follow an incorrect assumption of that distribution.

To solve this problem, in this paper, we propose a generative adversarial model architecture using InfoGAN to try to learn the data distributions and generate synthetic samples of ECG beats. The idea is to balance the dataset with sintentic data generated by an InfoGAN. We call this method adversarial oversampling. Moreover, we compared our proposed method with three other traditional oversampling methods: RandomOversampler, SMOTE, and ADASYN.

The MIT-BIH arrhythmia database was used to perform experiments in order to verify if a CNN training using the proposed method outperforms the ones trained using standard oversampling approaches.

\begin{figure*}
\centering
\includegraphics[width=0.9\textwidth]{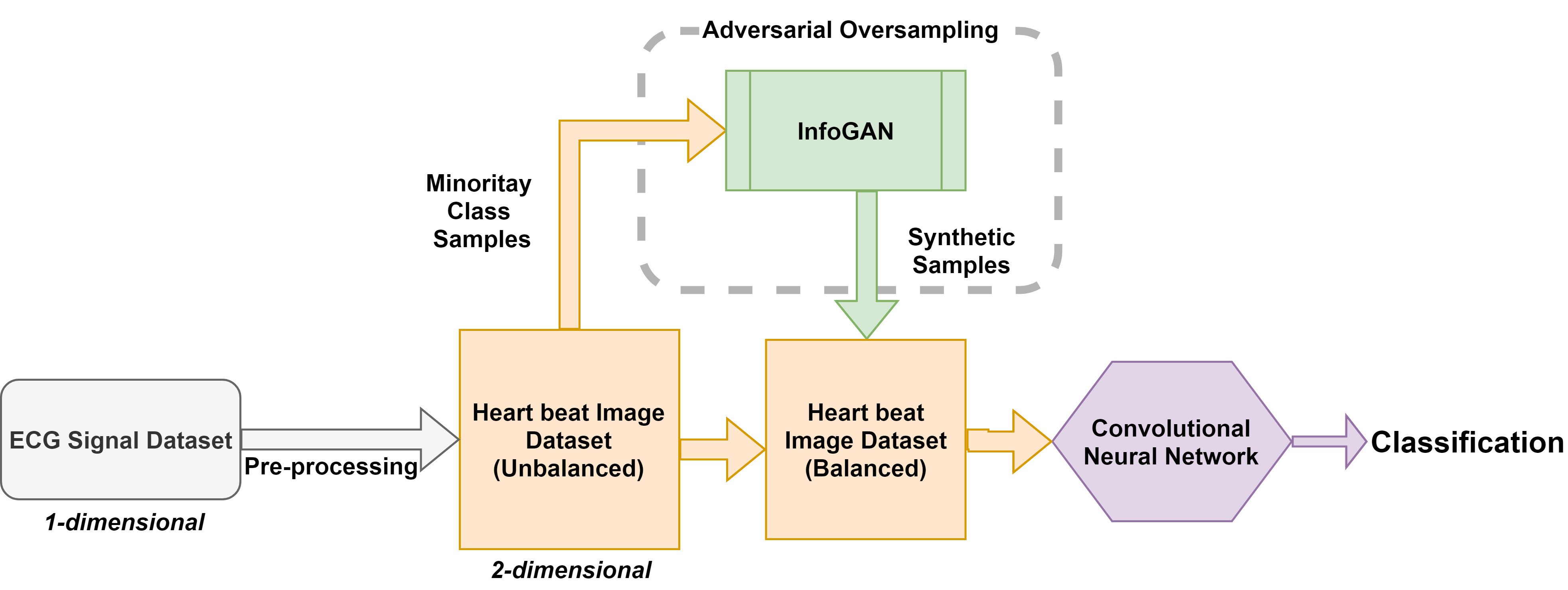}
\caption{Pipeline of the proposed model: Pre-processing (converts signals into 2D images of single beats); InfoGAN (generates synthetic samples to balance the dataset); CNN (classifies the database balanced).}
\label{fig:verticalcell}
\end{figure*}

In section II, we described related work. Section III presents the proposed method. Additionally, the experiments and results are reported in Section IV and Section V, respectively. Section VI presents final remarks and future works.

\section{Background}

This section describes the subjects that served as a basis for developing this work. First, we present the standard oversampling methods used in this paper. After that, we describe Generative Adversarial Networks (GAN) in generic terms.

\subsection{Oversampling Methods}

The present work makes use of synthetic samples generated by InfoGAN to perform a dataset balancing. To investigate the effectiveness of this approach, we compare to three other classical methods for data balancing: RandomOversampler, SMOTE, and ADASYN. These methods are widespread in the literature and are widely used to deal with unbalanced datasets \cite{smote_1} \cite{smote_2} \cite{smote_3} \cite{adasyn_1} \cite{adasyn_2} \cite{adasyn_3}. All methods are available in the Imblearn library \cite{imblearn_paper}.

\subsubsection{Random OverSampler} 

It consists of the most straightforward approach to performing Oversampling. The method suggests choosing samples from the minority class randomly and with replacement.

\subsubsection{SMOTE: Synthetic Minority Over-sampling Technique}

To oversample with SMOTE \cite{13}, is took a sample from the dataset, and consider its k nearest neighbors (in feature space). To create a synthetic data point, take the vector between one of those k neighbors, and the current data point. Basically, synthesizes new instances between existing (real) minority instances. SMOTE draws lines between existing minority samples and then generate new, synthetic minority instances somewhere on these lines.  

\subsubsection{ADASYN: Adaptive Synthetic Sampling Approach for Imbalanced Learning}

The basic idea behind ADASYN \cite{14} is to consider the level of difficulty in learning different samples from minority class examples and use to weighted distribution for each class. The method generates synthetic data for minority class examples that are harder to learn compared to those minority examples that are easier to learn. That is, forcing the learning algorithm to focus on regions of difficult learning. The main idea of the ADASYN algorithm is to use a density distribution as a criterion for automatically deciding the needed number of synthetic samples to be generated for each example of the minority class. So, for each example of the minority class $x_i$,  $g_i$ synthetic examples are generated.

\subsection{Generative Adversarial Networks}

GANs \cite{8} have shown remarkable success as a framework for training models to produce realistic-looking data. GANs are composed of two networks which compete in other to improve their performance. Given a training set \textbf{X}, the \textit{Generator}, \textbf{G}(\textbf{x}), takes as input a noise vector and tries to produce sample data similar to real data presented in the training set.

A \textit{Discriminator} network, \textbf{D(x)}, is a binary classifier that tries to distinguish between the real data exhibited in the training set \textbf{X} and the fake data generated by the Generator. So, given a set of training data, GANs can learn to estimate the underlying probability distribution of the training data. 

Based on traditional GAN models, Xi Chen \textit{et al.} \cite{15} (2016) designed the infoGAN, which applies concepts from Information Theory to transform some of the noise terms into latent codes that have systematic and predictable effects on the outcome. InfoGAN performs its task by splitting the Generator input code into two parts: the traditional noise vector and a new “latent code” vector. The codes are then made meaningful by maximizing the Mutual Information between the code and the generator output. It allows convergence to happen much faster than in traditional GANs. 

\begin{figure*}
\centering
\includegraphics[width=0.9\textwidth]{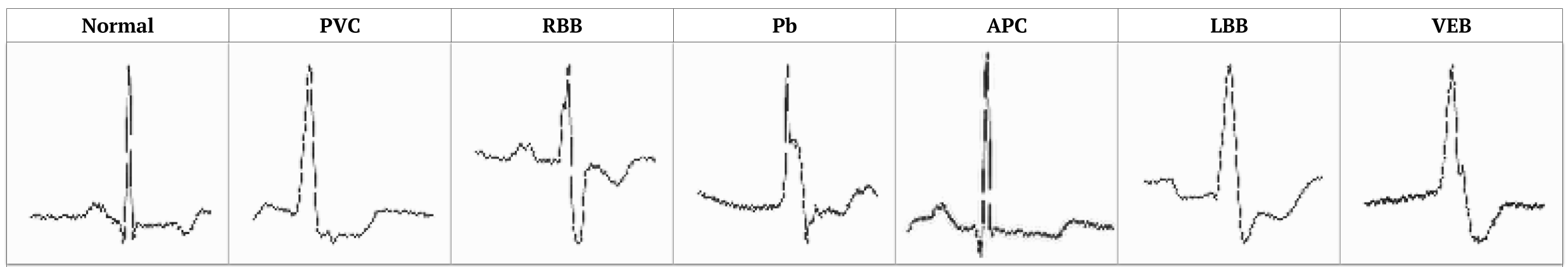}
\caption{Samples of heartbeat after the pre-processing step with their respective classes.}
\label{fig:data_imgs_plots}
\end{figure*}


Recently, the use of synthetic samples generated by GANs has been gaining strength with the results obtained in some works. Anthreas Antoniou \textit{et al.} (2018) \cite{9} presented the Data Augmentation Generative Adversarial Networks (DAGAN). It has been empirically shown that the samples generated by DAGAN generally yielded gains of accuracy, reaching improvements of up to 13\% in experiments. M. Frid-Adar \textit{et al.} (2018) \cite{10} used synthetic samples generated by a Deep Convolutional GAN (DCGAN) to perform Data Augmentation in order to improve liver lesion classification. It achieved an improvement of 7\% using synthetic augmentation over the classic augmentation methods. F. Bolelli \textit{et al.} (2018) \cite{11} used GANs in Skin Lesion Segmentation problem. The GANs was used to augment data in the image segmentation field, and a Convolutional Deconvolutional Neural Network (CDNN) to generate lesion segmentation mask from dermoscopic images.

Among the surveys, we have no find any approach to address the use of GAN to deal with unbalanced datasets involving health problems, therefore, more like the Oversampling we have is the use of GAN to Data Augmentation for a whole dataset.

\section{Heartbeat Anomaly Detection \\ using Adversarial Oversampling}

In this section, we describe the complete pipeline of the model proposed in this work. The Figure~\ref{fig:verticalcell} shows in a summarized way the flow of the solution. Three phases comprise the method: Pre-processing, Adversarial Oversampling (InfoGAN), and CNN classification. 

\begin{enumerate}
\item \textit{Pre-processing}: The complete long ECG records are preprocessed to extract all individual beats and produce heartbeat centered binary images;  
\item \textit{Adversarial Oversampling}: The InfoGAN is used to generate synthetic adversarial images for the minority heartbeat classes;
\item \textit{Trainning and Classification}: The new balanced dataset is used to training a 2D CNN to image classification.
\end{enumerate}

\subsection{Pre-processing}
 
After the acquisition of the signal, a pre-processing was performed to collect a range of 25 time-slices of the signal. Those images have the peak of the heartbeat in its central region. Therefore, each peak centralized image is classified individually. Then, we produced binary images with $112\!\times\!112$ resolution corresponding to the peaks. With this, the database that was previously composed by a long 1-dimensional ECG series is now composed of several 2-dimensional labeled images. The Figure~\ref{fig:data_imgs_plots} shows the data after all pre-processing.

\subsection{Adversarial Oversampling}

After a search of the existing GAN models, rapid experiments were carried out to evaluate the speed of convergence, robustness to mode collapse and visual quality of the synthetic generated images. We evaluated in details the traditional GAN, the Wasserstein GAN \cite{wgan_paper}, and InfoGAN. Finally, we chose the InfoGAN as the adversary model.

As we need to generate new images, the Generator (\textbf{G}) and Discriminator (\textbf{D}) are both composed by convolutional layers. The Generator also has upsampling and batch normalization layers. At the end of the model, we employed a \textit{Tanh} activation function to have activation between $-1$ and $1$.

Therefore, the \textbf{G} takes a standard normal random noise of size $64\!\times\!1$, which turns on a synthetic image at the end of the model. The \textbf{D} takes both fake and real images (not at same time) with size $112\!\times\!112$. The \textbf{D} is composed of 2-dimensional convolutional layers, dropout to minimize overfitting, batch normalization, Leaky ReLU \cite{12} as activation function, and Sigmoid function as the output of the model. The auxiliary \textbf{Q Net} is composed of Fully Connected layers and a Softmax function as output to give us the current label of the synthetic image.

\begin{figure*}
\centering
\subfloat[]{
\includegraphics[width=0.4\textwidth]{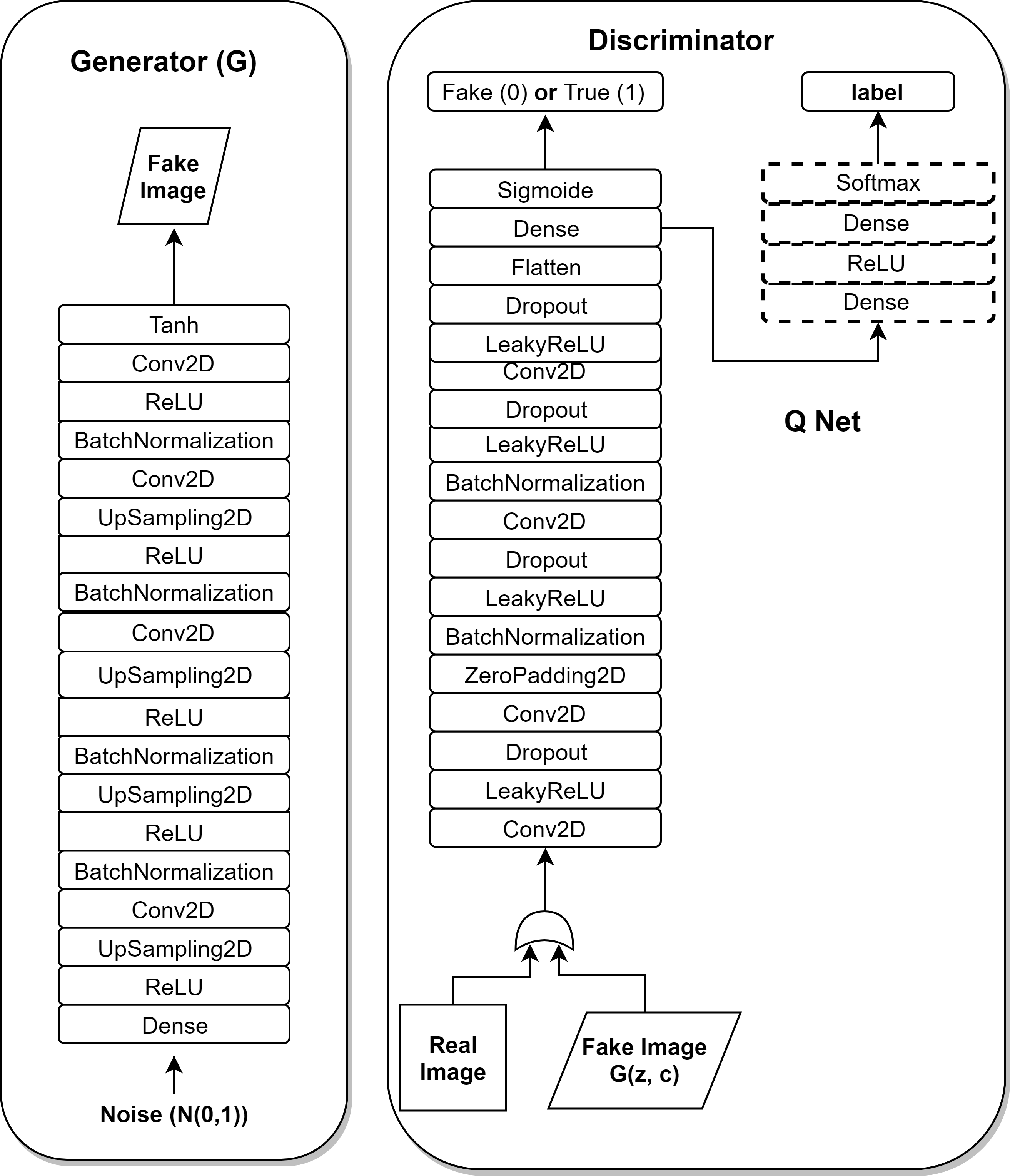}
}
\hskip 1.5cm
\subfloat[]{
\includegraphics[width=0.2\textwidth]{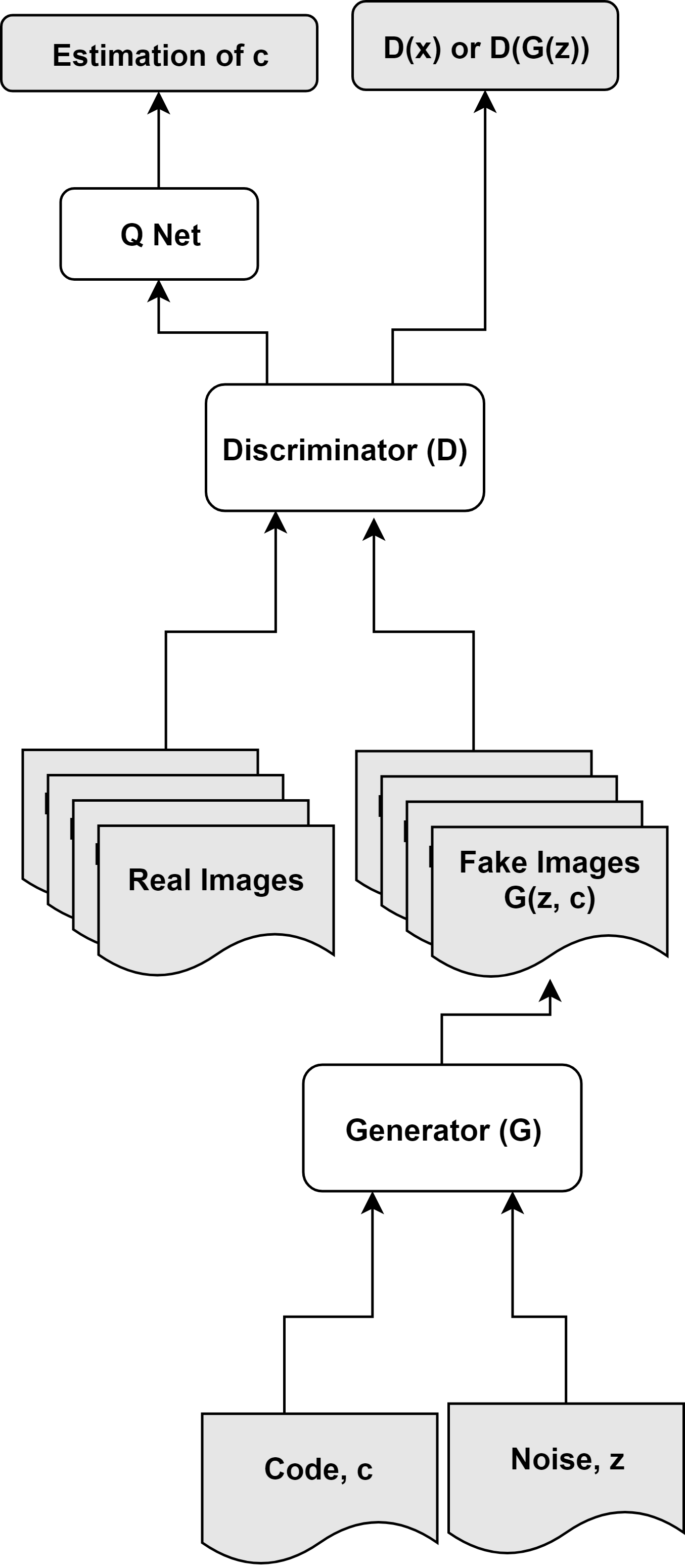}
}
\hskip 1.5cm
\subfloat[]{
\includegraphics[width=0.135\textwidth]{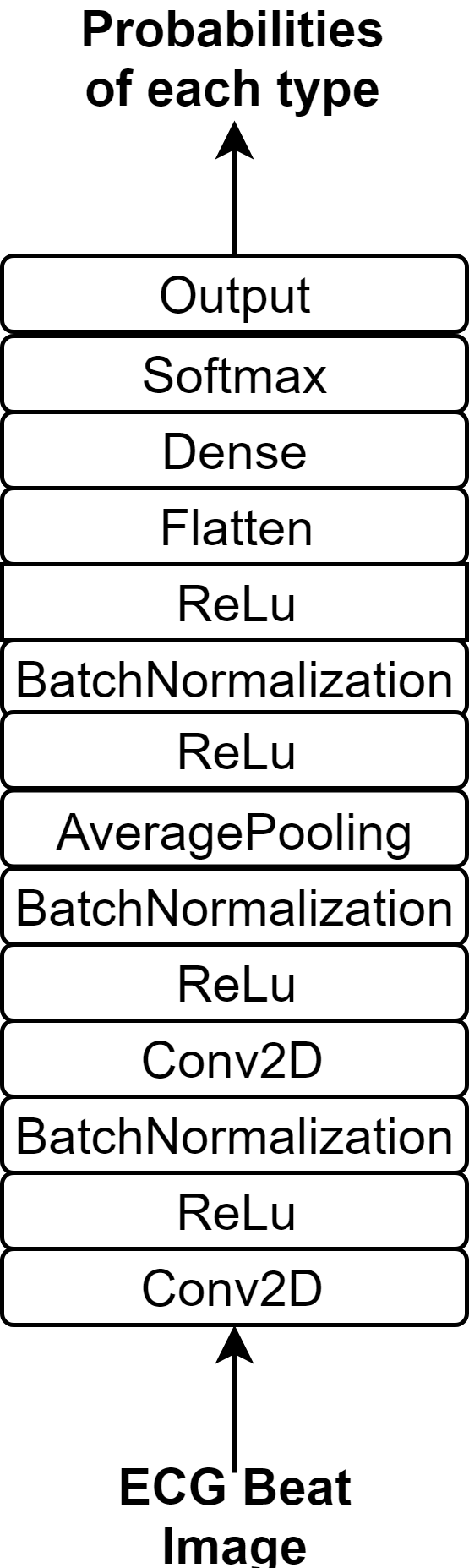}
}
\caption{(a) \textit{InfoGAN} Architecture. (b) \textit{InfoGAN} Training. (c) CNN Architecture.}
\label{fig:infoga_arch}
\end{figure*}

The Figure~\ref{fig:infoga_arch}(a) shows at left the architecture of the \textit{InfoGAN} Generator, and a detailed architecture of the Discriminator (\textbf{D}) and Auxiliary \textbf{Q Net} at right. 

To train the InfoGAN, we use batches of noise generated images as fake class and batches of real images as the true class. So, the traditional steps to train a conventional GAN is performed. We pass a batch of samples with some fake data, and a batch of samples with real data to the Discriminator to learn how to differentiate fake data from the real data. The InfoGAN training is shown in the Figure~\ref{fig:infoga_arch}(b).

After training some epochs of the Discriminator and the \textbf{Q Net}, we freeze \textbf{D} weights and train only the Generator model at the combined model. Consequently, the Generator model iteratively learns how to generate synthetic images that follow the real distribution of the training images. The general InfoGAN objective function is given by the lower-bound approximation to the Mutual Information as follows:

\begingroup\makeatletter\def\f@size{9}\check@mathfonts
$$min(G,Q)max(D)V_{infoGAN}(D,G,Q) = V(D, G) - \lambda L_I(G, Q)$$
\endgroup

We adopted the following methodology to generate synthetic samples: Given training data samples from the minority classes VEB and APC, we train the InfoGAN for both samples. The training/generation of images was iterative. The max number of epochs was 100 thousand, and at every 500 epochs, we save a snapshot of the generator network model and some synthetic images. When the training was finished, we looked at all samples and chose snapshots that generated images with a realistic appearance.  Hence, we are not considering the loss, but looking only to the quality of generated images.

\subsection{Training and Classification}

A Convolutional Neural Network (CNN) was chosen as the classification model. The Figure~\ref{fig:infoga_arch}(c) presents the high-level architecture of the proposed CNN. The network takes as input a 2D dimensional array, which represents a sampled heartbeat image with $112\!\times\!112$ dimensions. Two convolutional layers compose the model. The kernels initialization follow the He Initialization approach \cite{5}. We used Rectified Linear Unit (ReLU) \cite{relupaper} as the activation function. 

\begin{equation}
ReLU(x) = max(0,x)
\end{equation}

To prevent overfitting, we employ some layers with the dropout training strategy \cite{6} and also batch normalization \cite{7} to accelerate training. As subsampling method, we use simple \textit{AveragePooling} layers. At the end of the network, we used a \textit{Softmax} layer to provide the output as probabilities of each heartbeat class. We use the Adam optimizer \cite{10} with the default parameters and interrupted the training when the loss on the validation set stopped to decrease.
¨
Given the data set \textbf{\textit{X}}, for a single sample of \textbf{\textit{$X_i$}} from \textbf{\textit{X}}, the network training consists in optimize the following cross-entropy objective function:

\begin{equation}
\label{eq:loss}
L(X_i,c) = -\sum_{c=1}^{M} y_{\left(o,c\right)} p_{\left(o,c\right)}
\end{equation}

In the previously equation, $M$ is the number of classes or rhythms that a signal can assume, $y$ is a binary indicator (0 or 1) if class label \textit{c} is the correct classification for observation (heartbeat) \textit{o}, and $p$ is the predicted probability observation \textit{o} is of class \textit{c}.

\begin{table*}[!t]
\centering
\label{table:best_cnn}
\caption{F1 score for each oversampling method}
\begin{tabular}{@{}lccccc@{}}
\toprule
Class & \textbf{Original} & \textbf{Adversarial Oversampling} & \textbf{ADASYN} & \textbf{SMOTE} & \textbf{Random Oversampler}\\
\midrule
\textbf{APC} & $0.68\pm0.03$ & \bm{$0.80\pm0.03$} & $0.73\pm0.02$ & $0.71\pm0.02$ & $0.72\pm0.03$\\
\midrule
\textbf{Normal} & $0.98\pm0.01$ & $0.98\pm0.01$ & $0.97\pm0.02$ & $0.95\pm0.02$ & $0.94\pm0.01$\\
\midrule
\textbf{LBBB} & $0.96\pm0.02$ & $0.96\pm0.02$ & $0.95\pm0.01$ & $0.96\pm0.01$ & $0.93\pm0.01$\\
\midrule
\textbf{PAB} & $0.92\pm0.01$ & $0.92\pm0.01$ & $0.92\pm0.01$ & $0.91\pm0.01$ & $0.91\pm0.02$\\
\midrule
\textbf{PVC} & $0.93\pm0.01$ & $0.93\pm0.02$ & $0.91\pm0.02$ & $0.91\pm0.01$ & $0.89\pm0.02$\\
\midrule
\textbf{RBB} & $0.93\pm0.02$ & $0.94\pm0.01$ & $0.93\pm0.01$ & $0.92\pm0.01$ & $0.92\pm0.01$\\
\midrule
\textbf{VEB} & $0.43\pm0.02$ & \bm{$0.81\pm0.02$} & $0.61\pm0.03$ & $0.55\pm0.02$ & $0.49\pm0.02$\\
\midrule
Total & $0.83\pm0.02$ & \bm{$0.91\pm0.02$} & $0.86\pm0.02$ & $0.84\pm0.01$ & $0.83\pm0.02$\\
\bottomrule
\end{tabular}
\end{table*}

\section{Experiments}

This section describes the experiments that investigate if Adversarial Oversampling could be used to balance the training data set. That is, whether synthetic data generated by a GAN could improve the overall classification accuracy. We compare the performance of this Adversarial Oversampling with the RandomOversampler, SMOTE, and ADASYN.

In this experiments, we use the MIT-BIH Arrhythmia Database, which is a classical arrhythmia database. This dataset contains 48 half-hour excerpts of two-channel ambulatory ECG 360hz recordings obtained from 47 subjects studied by the BIH Arrhythmia Laboratory between 1975 and 1979. \cite{mitdb}.  From the MIT-BIH Arrhythmia dataset, we collect a number of samples from seven different classes as shown in Table~\ref{table:dataset_table}.

\begin{table}[!b]
\centering
\caption{MIT-BIH Arrhythmia Datasset Description}
\label{table:dataset_table}
\renewcommand{\arraystretch}{1.6}
\begin{tabular}{@{}lcc@{}}
\toprule
\multicolumn{2}{c}{\textbf{Heartbeat Type}} & \multicolumn{1}{c}{\textbf{\# Samples}} \\
\midrule
Atrial Premature  Contraction      & APC     & \textbf{243}                                     \\
\midrule
Normal                             & Normal  & 1079                                    \\
\midrule
Left Bundle Branch Block           & LBBB    & 1051                                    \\
\midrule
Paced Beat                         & PAB     & 895                                     \\
\midrule
Premature Ventricular Contraction  & PVC     & 1012                                    \\
\midrule
Right Bundle Branch                & RBB     & 1006                                    \\
\midrule
Ventricular Escape Beat            & VEB     & \textbf{106}                                     \\
\bottomrule
\end{tabular}
\end{table}

The CNN training consisted of 20 epochs and an early-stop if the loss in the validation set did not decrease considering the previous epoch. The experiments were repeated ten times, where 20\% the data was used as the test set, 10\% as the validation set, and 70\% to training.

Considering the methodology used to generate InfoGAN synthetic samples,  we split the original dataset between the training set and test set, and only the train set was used to InfoGAN training. Considering only the training set, we generate how many samples were necessary to reach a total of 1000 samples as the other classes have around this amount. In this way, the new training set became relatively balanced.

After generating the synthetic samples, we perform the Adversarial Oversampling iteratively added the samples to the network training and observed the F1 score for each class. Then we used the same training and test environment and compare the Adversarial Oversampling with the traditional oversampling methods. The classifier for each oversampling methods was the proposed CNN method.

\section{Results and Discussion}

In this section, the results of the proposed Adversarial Oversampling approach are compared to the ones provided by traditional oversampling methods. As shown in the second column in Table I, the CNN produced low performance in the original (unbalanced) dataset influenced by the F1 score for the minority classes VEB and APC.

Consequently, this experiment aims to minimize this problem by generating high-quality synthetic samples to minority classes to get a more balanced train set. Although it had a slightly smaller amount, the PAB class was not chosen to be oversampled since it did not influence the performance of the overall classification.

\begin{figure}[!b]
\centering
\includegraphics[width=0.45\textwidth]{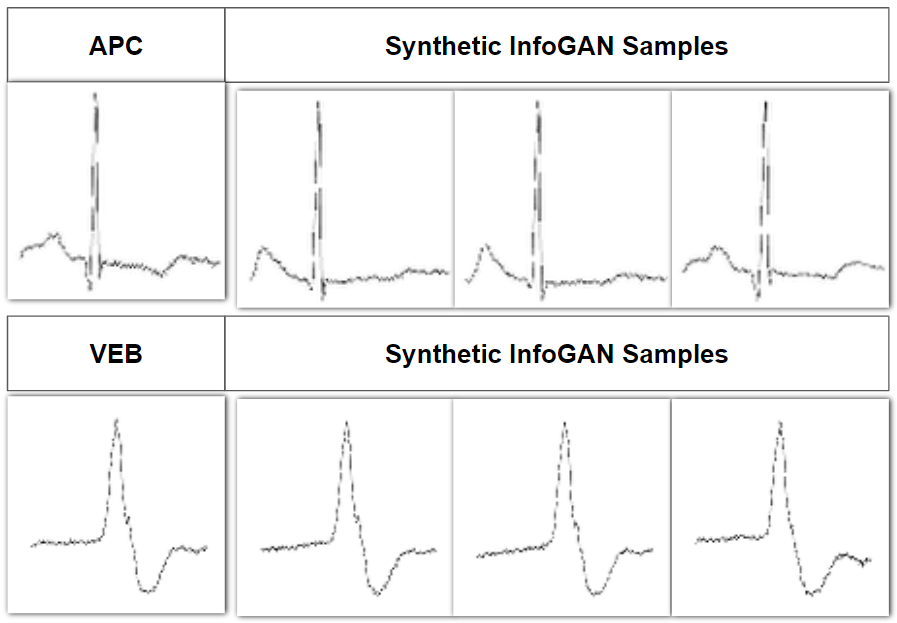}
\caption{Adversarial samples from VEB and APC classes.}
\label{fig:generated}
\end{figure}

The Figure~\ref{fig:generated} shows three adversarial samples for the classes APC and VEB. On the left side, we have two images representing the original samples of the mentioned clasess.

To observe whether Adversarial Oversampling could be used to perform the dataset balancing, the F1 score for each class was observed as synthetic samples were inserted. The objective was to observe if the APC and VEB minority classes would perform better in the classification, without decreasing the performance of the other classes.

\begin{figure*}[!t]
\centering
\subfloat[]{
\includegraphics[width=0.45\textwidth,height=160pt]{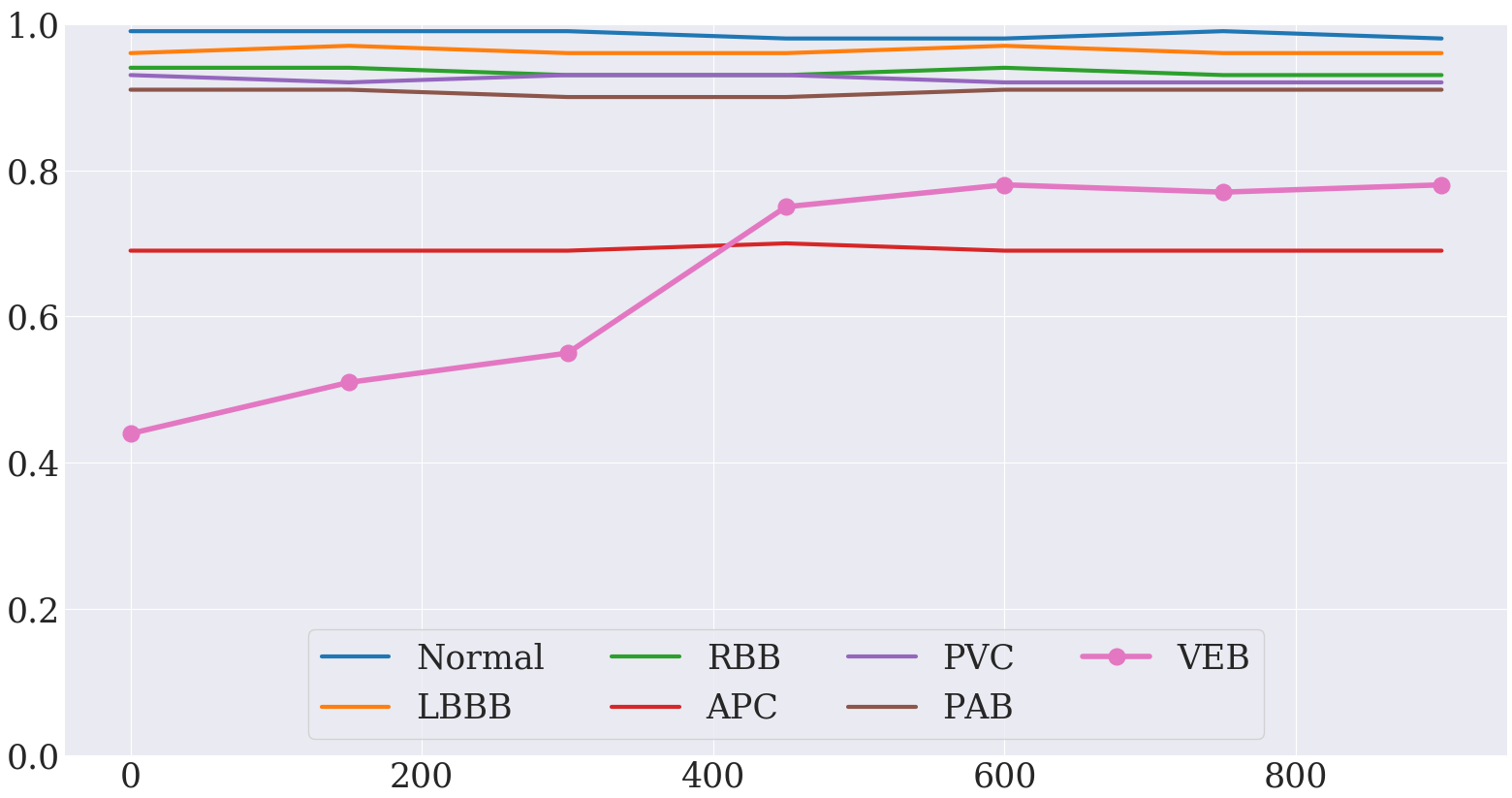}
\label{fig:veb_samples}
}
\hskip 20pt
\subfloat[]{
\includegraphics[width=0.45\textwidth,height=160pt]{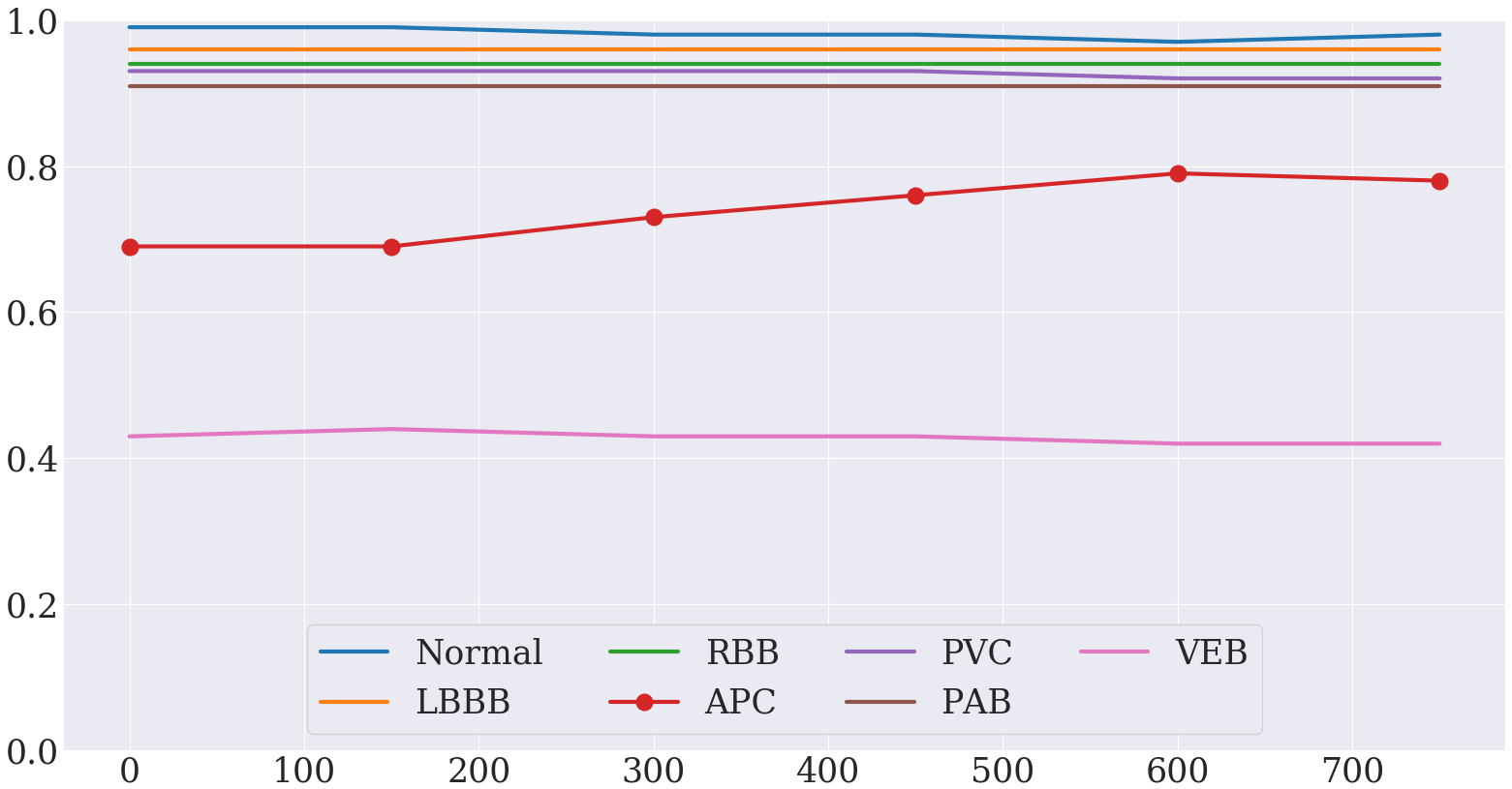}
\label{fig:apc_samples}
}
\caption{Mean of F1-score for each class when (a) VEB and (b) APC synthetic samples are used.}
\label{fig:samples}
\end{figure*}

The Figure~\ref{fig:veb_samples} and the Figure~\ref{fig:apc_samples} show the CNN performance for each class when 150 and 100 samples are added until reaching the total number of 1000. Both graphs contain a mean F1 score of 10 executions.

After observing the variation in CNN performance as synthetic samples are inserted into the training set, we tested the CNN using the best amount of synthetic samples for each of both classes. As shown in Figure~\ref{fig:veb_samples} and Figure~\ref{fig:apc_samples}, the best performance of CNN was observed using 600 synthetic samples of the APC class and 600 synthetic samples of the VEB class. The Table I shows the F1-score for each class using Adversarial Oversampling compared to the RandomOverSampler, ADASYN, SMOTE oversampling methods.

As we can see, the results favor the hypothesis that the Adversarial Oversampling could be used to perform the dataset balancing taking into account the performance observed in Table I, which was higher than obtained by oversampling using RandomOversampler, SMOTE and ADASYN for the two minority classes, APC and VEB.

One possible explanation for this may be the fact that InfoGAN generates new samples, not just repeats them as in the case of RandomOversampler. Besides, it can generate samples that follow the original distribution of the classes, reducing the occurrence of class noise, which is a common risk in traditional Oversampling methods.

\section{Conclusion}

When using deep learning for signal analysis, such as ECG, through convolutional neural networks, convolutions with one dimension consider the signal as a time series. However, according to the obtained results, we can observe that the bi-dimensional approach used in our method obtained an excellent performance. Only the minority classes did not have a satisfactory classification.

Concerning the performance of the \textit{InfoGAN} to generate synthetic samples, we show that these samples have a variation within the class and maintain the original characteristics of the signal, which is fundamental. So, we can see that and Adversarial Oversampling can be used to perform an dataset balancing, generating new synthetics samples of the minority classes on the dataset and that the generated and selected samples were able to decrease the effects of the unbalance data significantly. The proposed approach improved the performance of CNN in the classification of VEB and APC types.  Besides, for the used dataset the Adversarial Oversampling proposed overcomes traditional methods such as RandomOversampler, SMOTE, and ADASYN.

As future works we suggest the elaboration of a more automatic form of generating and selection samples, possibly observing only the loss, without the necessity of human interference to select good synthetic samples to use. Also, we suggest evaluating the performance of new approaches such as Unrolled GAN \cite{unrolledGAN}. The base code for this work is available publicly online (\url{https://github.com/JeffersonLPLima/adversarial\_oversampling}).

\newpage

\section*{Acknowledgment}

We would like to thank CAPES and CNPq (Brazilian research agencies) for the financial support. In addition the E.life Brazil for making the infrastructure available for conducting the experiments.

\bibliographystyle{ieeetr}
\bibliography{ref}

\end{document}